# Regularized Pel-Recursive Motion Estimation Using Generalized Cross-Validation and Spatial Adaptation

**Vania Estrela** [1], **Luís A. Rivera**[1], **Paulo C. Beggio**[1], **Ricardo T. Lopes**[2]

[1]*Universidade Estadual do Norte Fluminense, LCMAT-CCT, Grupo de Computação Científica,*
*Av. Alberto Lamego 2000, Campos dos Goytacazes, RJ, Brasil, CEP 28013-600*
{vaniave, rivera, beggio}@uenf.br

[2]*Laboratório de Instrumentação Nuclear EE/COPPE/UFRJ, RJ, Brasil, CEP 21945-970*
ricardo@lin.ufrj.br

## Abstract

*The computation of 2-D optical flow by means of regularized pel-recursive algorithms raises a host of issues, which include the treatment of outliers, motion discontinuities and occlusion among other problems. We propose a new approach which allows us to deal with these issues within a common framework. Our approach is based on the use of a technique called Generalized Cross-Validation to estimate the best regularization scheme for a given pixel. In our model, the regularization parameter is a matrix whose entries can account for diverse sources of error. The estimation of the motion vectors takes into consideration local properties of the image following a spatially adaptive approach where each moving pixel is supposed to have its own regularization matrix. Preliminary experiments indicate that this approach provides robust estimates of the optical flow.*

## 1. Introduction

Motion estimation is very important in multimedia video processing applications. For example, in video coding, the estimated motion is used to reduce the transmission bandwidth. The evolution of an image sequence motion field can also help other image processing tasks in multimedia applications such as analysis, recognition, tracking, restoration, collision avoidance and segmentation of objects [6].

In coding applications, a block-based approach [7] is often used for interpolation of lost information between key frames. The fixed rectangular partitioning of the image used by some block-based approaches often separates visually meaningful image features. If the components of an important feature are assigned different motion vectors, then the interpolated image will suffer from annoying artifacts. Pel-recursive schemes [2,3,6] can theoretically overcome some of the limitations associated with blocks by assigning a unique motion vector to each pixel. Intermediate frames are then constructed by resampling the image at locations determined by linear interpolation of the motion vectors. The pel-recursive approach can also manage motion with sub-pixel accuracy. However, its original formulation was deterministic. The update of the motion estimate was based on the minimization of the displaced frame difference (DFD) at a pixel. In the absence of additional assumptions about the pixel motion, this estimation problem becomes "ill-posed" because of the following problems: a) occlusion; b) the solution to the 2D motion estimation problem is not unique (aperture problem); and c) the solution does not continuously depend on the data due to the fact that motion estimation is highly sensitive to the presence of observation noise in video images.

We propose to solve optical flow (OF) problems by means of the Generalized Cross-Validation (GCV) framework, introducing a more complex regularization matrix $\Lambda$. Such approach accounts better for the statistical properties of the errors present in the scenes than the solution proposed by Biemond [1] where a scalar regularization parameter was used.

We organized this work as follows. Section 2 provides some necessary background on the pel-recursive motion estimation problem. Section 3 introduces our spatially adaptive approach. Section 4 describes the ordinary cross-validation. Section 5 deals with the GCV technique. Section 6 defines the metrics used to evaluate our results. Section 7 describes the experiments used to access the performance of our proposed algorithm. Finally, Section 8 has the conclusions.

## 2. Pel-Recursive Displacement Estimation

### 2.1. Problem Characterization

The displacement of each picture element in each frame forms the displacement vector field (DVF) and its estimation can be done using at least two successive frames. The DVF is the 2D motion resulting from the apparent motion of the image brightness (OF). A vector is assigned to each point in the image.

A pixel belongs to a moving area if its intensity has changed between consecutive frames. Hence, our goal is to find the corresponding intensity value $I_k(\mathbf{r})$ of the k-th frame at location $\mathbf{r} = [x, y]^T$, and $\mathbf{d}(\mathbf{r}) = [d_x, d_y]^T$ the corresponding (true) displacement vector (DV) at the working point $\mathbf{r}$ in the current frame. Pel-recursive algorithms minimize the DFD function in a small area containing the working point assuming constant image intensity along the motion trajectory. The DFD is defined by

$$\Delta(\mathbf{r}; \mathbf{d}(\mathbf{r})) = I_k(\mathbf{r}) - I_{k-1}(\mathbf{r}-\mathbf{d}(\mathbf{r})) \quad (1)$$

and the perfect registration of frames will result in $I_k(\mathbf{r})=I_{k-1}(\mathbf{r}-\mathbf{d}(\mathbf{r}))$. The DFD represents the error due to the nonlinear temporal prediction of the intensity field through the DV. The relationship between the DVF and the intensity field is nonlinear. An estimate of $\mathbf{d}(\mathbf{r})$, is obtained by directly minimizing $\Delta(\mathbf{r},\mathbf{d}(\mathbf{r}))$ or by determining a linear relationship between these two variables through some model. This is accomplished by using the Taylor series expansion of $I_{k-1}(\mathbf{r}-\mathbf{d}(\mathbf{r}))$ about the location $(\mathbf{r}-\mathbf{d}^i(\mathbf{r}))$, where $\mathbf{d}^i(\mathbf{r})$ represents a prediction of $\mathbf{d}(\mathbf{r})$ in i-th step. This results in

$$\Delta(\mathbf{r}, \mathbf{r}-\mathbf{d}^i(\mathbf{r})) = -\mathbf{u}^T \nabla I_{k-1}(\mathbf{r}-\mathbf{d}^i(\mathbf{r})) + e(\mathbf{r}, \mathbf{d}(\mathbf{r})), \quad (2)$$

where the displacement update vector $\mathbf{u}=[u_x, u_y]^T = \mathbf{d}(\mathbf{r}) - \mathbf{d}^i(\mathbf{r})$, $e(\mathbf{r}, \mathbf{d}(\mathbf{r}))$ represents the error resulting from the truncation of the higher order terms (linearization error) and $\nabla=[\partial/\partial_x, \partial/\partial_y]^T$ represents the spatial gradient operator. Applying (2) to all points in a neighborhood $\mathcal{R}$ gives

$$\mathbf{z} = \mathbf{G}\mathbf{u} + \mathbf{n}, \quad (3)$$

where the temporal gradients $\Delta(\mathbf{r}, \mathbf{r}-\mathbf{d}^i(\mathbf{r}))$ have been stacked to form the N×1 observation vector $\mathbf{z}$ containing DFD information on all the pixels in a neighborhood $\mathcal{R}$, the N×2 matrix $\mathbf{G}$ is obtained by stacking the spatial gradient operators at each observation, and the error terms have formed the N×1 noise vector $\mathbf{n}$ which is assumed Gaussian with $\mathbf{n} \sim N(0, \sigma_n^2 \mathbf{I})$. Each row of $\mathbf{G}$ has entries $[g_{xi}, g_{yi}]^T$, with $i = 1, \ldots, N$. The spatial gradients of $I_{k-1}$ are calculated through a bilinear interpolation scheme [2].

## 2.2. Regularized Least Squares Estimation

The pel-recursive estimator for each pixel located at position $\mathbf{r}$ of a frame can be written as

$$\mathbf{d}^{i+1}(\mathbf{r}) = \mathbf{d}^i(\mathbf{r}) + \mathbf{u}^i(\mathbf{r}), \quad (4)$$

where $\mathbf{u}^i(\mathbf{r})$ is the current motion update vector obtained through a motion estimation procedure that attempts to solve (3), $\mathbf{d}^i(\mathbf{r})$ is the DV at iteration i and $\mathbf{d}^{i+1}(\mathbf{r})$ is the corrected DV. The regularized minimum norm solution to the previous expression, that is

$$\hat{\mathbf{u}}(\Lambda) = \hat{\mathbf{u}}_{RLS}(\Lambda) = (\mathbf{G}^T\mathbf{G} + \Lambda)^{-1}\mathbf{G}^T\mathbf{z}, \quad (5)$$

is also known as regularized least square (RLS) solution.

In order to improve the RLS estimate of the motion update vector, we propose a strategy which takes into consideration the local properties of the image. It is described in the next section.

## 3. Spatially Adaptive Neighborhoods

Aiming to improve the estimates given by the pel-recursive algorithm, we introduced an adaptive scheme for determining the optimal shape of the neighborhood of pixels with the same DV used to generate the overdetermined system of equations given by (3). More specifically, the masks in Figure 1 show the geometries of the neighborhoods used.

Errors can be caused by the basic underlying assumption of uniform motion inside $\mathcal{R}$ (the smoothness constraint), by not grouping pixels adequately, and by the way gradient vectors are estimated, among other things. Since it is known that in a noiseless image not containing pixels with constant intensity, most errors, when estimating motion, occur close to motion boundaries, we propose a hypothesis testing (HT) approach to determine the best neighborhood shape for a given pixel. We pick up the neighborhood from the finite set of templates shown in Figure 1, according to the smallest $|DFD|$ criterion, in an attempt to adapt the model to local features associated to motion boundaries.

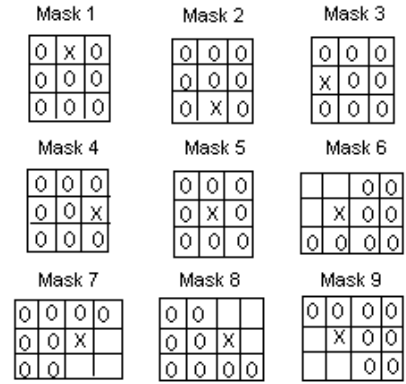

X Current pixel    O Neighboring pixel

Figure 1: Neighborhood geometries

## 4. Ordinary Cross-Validation (OCV)

The degree of smoothing of the solution $\hat{\mathbf{u}}(\Lambda)$, in (5), is dictated by the regularization matrix $\Lambda$. Cross-validation has been proven to be a very effective method of estimating the regularization parameters [4, 5, 8], which in our work are the entries of $\Lambda$, without any prior knowledge on the noise statistics.

The main advantage of the OCV is its systematic way of determining the regularization parameter directly from the observed data. However, it presents the following drawbacks:
1. It uses a noisy performance measure, Mean Squared Error (MSE). This means that since we are looking at the average value of the MSE over several observation sets re-sampled from the original $\mathbf{z}$, we can only guarantee the OCV estimator of $\Lambda$ is going to be a good predictor when $N \gg 1$.
2. It treats all data sets equally. In terms of image processing, we expect close neighbors of the current pixel to behave more similarly to it (in most of the cases) than pixels that are more distant from it. Of course, this is not the case with motion boundaries, occlusion and transparency.

## 5. The Generalized Cross-Validation (GCV)

The OCV does not provide good estimates of $\Lambda$ [5, 8]. A modified method called GCV function gives more satisfactory results. GCV is a weighted version of the OCV, and it is given by

$$GCV(\Lambda) = \frac{1}{N}\sum_{i=1}^{N}[z_i - \hat{z}_i]^2 w_i(\Lambda), \qquad (6)$$

where the weights $w_i$ are defined as follows:

$$w_i(\Lambda) = \left\{\frac{[1-a_{ii}(\Lambda)]}{\left(1-\frac{1}{N}Tr[\mathbf{A}(\Lambda)]\right)}\right\}^2, \text{ and} \qquad (7a)$$

$$\mathbf{A}(\Lambda) = \mathbf{G}(\mathbf{G}^T\mathbf{G}+\Lambda)^{-1}\mathbf{G}^T, \qquad (7b)$$

with $a_{ii}(\Lambda)$ being the diagonal entries of matrix $\mathbf{A}(\Lambda)$ as defined in (7b). The main shortcoming of OCV is the fact that OCV is not invariant to orthonormal transformations. In other words, if data $\mathbf{z}' = \Gamma\mathbf{z}$ is available, where $\Gamma$ is an $N \times N$ orthonormal matrix, and $\mathbf{z}'$ is the observation vector corresponding to the linear model given by

$$\mathbf{z}' = \mathbf{G}'\mathbf{u}'+\mathbf{n}' = \Gamma\{\mathbf{Gu}+\mathbf{n}\}. \qquad (8)$$

Therefore, the *OCV*, and, consequently the regularization matrix $\Lambda$, depends on $\Gamma$. *GCV* on the other hand is independent of $\Gamma$. Thus, the $GCV(\Lambda)$ is a better criterion for estimating the regularization parameters [5, 8]. So, (6) can also take the form

$$GCV(\Lambda) = \frac{1}{N}\frac{\left\|[\mathbf{I}-\mathbf{A}(\Lambda)]\mathbf{z}\right\|^2}{\left[\frac{1}{N}Tr\{\mathbf{I}-\mathbf{A}(\Lambda)\}\right]^2}. \qquad (9)$$

### 5.1. Regularization Parameter Determination

The *GCV* function for the observation model in (3) is given in closed form by (9). Let us call $\hat{\mathbf{u}}_{GCV}$ the solution for (3) when an optimum parameter set (the entries of the regularization matrix) $\Lambda_{GCV}$ is found by means of the GCV. Then, (5) becomes

$$\hat{\mathbf{u}}_{GCV} = (\mathbf{G}^T\mathbf{G}+\Lambda_{GCV})^{-1}\mathbf{G}^T\mathbf{z} \qquad (10)$$

### 5.2. The GCV-Based Estimation Algorithm

For each pixel located at $\mathbf{r} = (x, y)$ the GCV-based algorithm is described by the following steps:
1) Initialize the system: $\mathbf{d}^0(\mathbf{r})$, $m \leftarrow 0$ ($m$ = mask counter), and $i \leftarrow 0$ ($i$ = iteration counter).
2) If $|DFD| < T$, then stop. T is a threshold for $|DFD|$.
3) Calculate $\mathbf{G}^i$ and $\mathbf{z}^i$ for the current mask and current initial estimate.
4) Calculate $\Lambda^i$ by minimizing the expression

$$GCV(\Lambda^i) = \frac{1}{N}\frac{\left\|[\mathbf{I}-\mathbf{A}(\Lambda^i)]\mathbf{z}^i\right\|^2}{\left[\frac{1}{N}Tr\{\mathbf{I}-\mathbf{A}(\Lambda^i)\}\right]^2}, \qquad (11)$$

where $\mathbf{A}(\Lambda^i) = \mathbf{G}^i\left[(\mathbf{G}^i)^T\mathbf{G}^i+\Lambda^i\right]^{-1}(\mathbf{G}^i)^T$. (12)

5) Calculate the current update vector:
$$\mathbf{u}^i = \left[(\mathbf{G}^i)^T\mathbf{G}^i+\Lambda^i\right]^{-1}(\mathbf{G}^i)^T\mathbf{z}^i \qquad (13)$$

6) Calculate the new DV: $\mathbf{d}^{i+1}(\mathbf{r}) = \mathbf{d}^i(\mathbf{r}) + \mathbf{u}^i(\mathbf{r})$. (14)
7) For the current mask $m$:
If $\|\mathbf{d}^{i+1}(\mathbf{r})-\mathbf{d}^i(\mathbf{r})\| \leq \varepsilon$ and $|DFD| < T$, then stop.
If $i < (I - 1)$, where I is the maximum number of iterations allowed, then go to step 3 with $i \leftarrow i+1$ and use $\mathbf{d}^i(\mathbf{r}) \leftarrow \mathbf{d}^{i+1}(\mathbf{r})$ as the new initial estimate.
Otherwise, try another mask: $m \leftarrow m+1$. If all masks where used and no DV was found, then set $\mathbf{d}^{i+1}(\mathbf{r}) = 0$.

## 6. Metrics to Evaluate the Experiments

This work assesses the motion field quality through the use of the four metrics [2, 3] as described below.

### 6.1. Mean Squared Error (MSE)

Since the MSE provides an indication of the degree of correspondence between the estimates and the true value of the motion vectors, we can apply this measure to two consecutive frames of a sequence with known motion. We can evaluate the MSE in the horizontal ($MSE_x$) and in the vertical ($MSE_y$) directions as follows:

$$MSE_x = \frac{1}{RC}\sum_{\mathbf{r}\in\mathbf{S}}[d_x(\mathbf{r})-\hat{d}_x(\mathbf{r})]^2 \text{, and} \quad (15)$$

$$MSE_y = \frac{1}{RC}\sum_{\mathbf{r}\in\mathbf{S}}[d_y(\mathbf{r})-\hat{d}_y(\mathbf{r})]^2, \quad (16)$$

where $S$ is the entire frame, $\mathbf{r}$ represents the pixel coordinates, $R$ and $C$ are, respectively, the number of rows and columns in a frame, $\mathbf{d(r)}=(d_x(\mathbf{r}), d_y(\mathbf{r}))$ is the true DV at $\mathbf{r}$, and $\hat{\mathbf{d}}(\mathbf{r}) = (\hat{d}_x(\mathbf{r}),\hat{d}_y(\mathbf{r}))$ its estimation.

### 6.2. Bias

The bias gives an idea of the degree of correspondence between the estimated motion field and the original optical flow. It is defined as the average of the difference between the true DV's and their predictions, for all pixels inside a frame S, and it is defined along the x and y directions as

$$bias_x = \frac{1}{RC}\sum_{\mathbf{r}\in\mathbf{S}}[d_x(\mathbf{r})-\hat{d}_x(\mathbf{r})] \text{, and} \quad (17)$$

$$bias_y = \frac{1}{RC}\sum_{\mathbf{r}\in\mathbf{S}}[d_y(\mathbf{r})-\hat{d}_y(\mathbf{r})]. \quad (18)$$

### 6.3. Mean-Squared Displaced Frame Difference

This metric evaluates the behavior of the average of the squared displaced frame difference ($\overline{DFD}^2$). It represents an assessment of the evolution of the temporal gradient as the scene evolves by looking at the squared difference between the current intensity $I_k(\mathbf{r})$ and its predicted value $I_{k-1}(\mathbf{r-d(r)})$. Ideally, the $\overline{DFD}^2$ should be zero, which means that all motion was identified correctly ($I_k(\mathbf{r})= I_{k-1}(\mathbf{r-d(r)})$ for all $\mathbf{r}$'s). In practice, we want the $\overline{DFD}^2$ to be as low as possible. Its is defined as

$$\overline{DFD}^2 = \frac{\sum_{k=2}^{K}\sum_{\mathbf{r}\in\mathbf{S}}[I_k(\mathbf{r})-I_{k-1}(\mathbf{r-d(r)})]^2}{RC(K-1)}, \quad (19)$$

where $K$ is the length of the image sequence.

### 6.4. Improvement in Motion Compensation

The average improvement in motion compensation $\overline{IMC}(dB)$ between two consecutive frames is given by

$$\overline{IMC}_k(dB) = 10\log_{10}\left\{\frac{\sum_{\mathbf{r}\in\mathbf{S}}[I_k(\mathbf{r})-I_{k-1}(\mathbf{r})]^2}{\sum_{\mathbf{r}\in\mathbf{S}}[I_k(\mathbf{r})-I_{k-1}(\mathbf{r-d(r)})]^2}\right\} \quad (20)$$

where S is the frame being currently analyzed. It shows the ratio in decibel (dB) between the mean-squared frame difference ($\overline{FD}^2$) defined by

$$\overline{FD}^2 = \frac{\sum_{\mathbf{r}\in\mathbf{S}}[I_k(\mathbf{r})-I_{k-1}(\mathbf{r})]^2}{RC}, \quad (21)$$

and the $\overline{DFD}^2$ between frames $k$ and $(k-1)$.

As far as the use of the this metric goes, we chose to apply it to a sequence of $K$ frames, resulting in the following equation for the average improvement in motion compensation:

$$\overline{IMC}(dB) = 10\log_{10}\left\{\frac{\sum_{k=2}^{K}\sum_{\mathbf{r}\in\mathbf{S}}[I_k(\mathbf{r})-I_{k-1}(\mathbf{r})]^2}{\sum_{k=2}^{K}\sum_{\mathbf{r}\in\mathbf{S}}\left[I_k(\mathbf{r})-I_{k-1}(\mathbf{r-d(r)})\right]^2}\right\}. \quad (22)$$

When it comes to motion estimation, we seek algorithms that have high values of $\overline{IMC}(dB)$. If we could detect motion without any error, then the denominator of the previous expression would be zero (perfect registration of motion) and we would have $\overline{IMC}(dB) = \infty$.

## 7. Implementation

In this section, we present several experimental results that illustrate the effectiveness of the GCV approach and compare it with the Wiener filter [2, 3, 7] similar to the one in [1] given by

$$\hat{\mathbf{u}}_{Wiener} = \hat{\mathbf{u}}_{LMMSE} = (\mathbf{G}^T\mathbf{G}+\mu\mathbf{I})^{-1}\mathbf{G}^T\mathbf{z}, \quad (23)$$

where μ=50 was chosen for all pixels of the entire frame. All sequences are 144 x 176, 8-bit (QCIF format).

The algorithms were applied to three image sequences: one synthetically generated, with known motion; the "Mother and Daughter" (MD) and the "Foreman". For each sequence, two sets of experiments are analyzed: one for the noiseless case and the other for

a sequence whose frames are corrupted by a signal-to-noise-ratio (SNR) equal to 20 dB. The SNR is defined as

$$SNR = 10\log_{10}\frac{\sigma^2}{\sigma_c^2}. \quad (24)$$

where $\sigma^2$ is the variance of the original image and $\sigma_c^2$ is the variance of the noise corrupted image [8].

### 7.1. Programs and Experiments

The following GCV-based programs were developed:
a) **LSCRV:** $\lambda$ is a scalar; a non-causal 3x3 mask, centered at the pixel being analyzed.
b) **LSCRVB:** $\lambda$ is a scalar; we tried all nine masks.
c) **LSCRV1:** $\Lambda = \Lambda_{GCV} = \text{diag}(\lambda_1, \lambda_2)$, where $\lambda_1$ and $\lambda_2$ are scalars, is a matrix; a non-causal 3x3 mask centered at the pixel being analyzed.
d) **LSCRV2:** $\Lambda$ is a matrix; we tried all nine masks.

Results from the proposed algorithms are compared to the ones obtained with the Wiener (LMMSE) filter from (23) in the subsequent experiments.

**Experiment 1.** In this sequence, there is a moving rectangle immersed in a moving background. In order to create textures for the rectangle and its background (otherwise motion detection would not be possible), the following auto-regressive model was used:

$I(m,n) = \frac{1}{3}[I(m,n-1)+I(m-1,n)+I(m-1,n-1)]+ n_i(m,n),$
(25)

where i=1,2. For the background (i=1), $n_1$ is a Gaussian random variable with mean $\mu_1 = 50$ and variance $\sigma_1^2 = 49$. The rectangle (i=2) was generated with $\mu_2 = 100$ and variance $\sigma_2^2 = 25$. All pixels from the background move to the right, and the displacement from frame 1 to frame 2 is $\mathbf{d}_b(\mathbf{r})=(d_{bx}(\mathbf{r}),d_{by}(\mathbf{r}))=(2,0)$. The rectangle moves in a diagonal fashion from frame 1 to 2 with $\mathbf{d}_r(\mathbf{r})=(d_{rx}(\mathbf{r}),d_{ry}(\mathbf{r}))=(1,2)$.

Table 1 shows the values for the MSE, bias, $\overline{IMC}$ (dB) and $\overline{DFD}^2$ for the estimated optical flow using the Wiener filter and the four programs mentioned previously when no noise is present.

All the algorithms employing the GCV show improvement in terms of the metrics used. When we compare LSCRV with the Wiener filter, we see that with a regularization matrix of the form $\Lambda=\lambda\mathbf{I}$, whose regularization parameter $\lambda$ is determined by means of the minimization of the GCV function and using the same 3x3 mask as the Wiener, the improvements are small (we discuss some of our findings about the drawbacks of the GCV at the end of this article). When we introduce the spatially adaptive approach with nine masks, but keeping the regularization parameter a scalar $\lambda$ (algorithm LSCRVB), the performance of the GCV increases. Now, when we compare the performance of the previous algorithms with the case where we have a more complex regularization matrix $\Lambda=\text{diag}\{\lambda_1, \lambda_2\}$, that is, the implementation LSCRV1, then get even more improvements, although we have a single mask. Finally, using both the spatially adaptive approach and $\Lambda=\text{diag}\{\lambda_1, \lambda_2\}$ we get the best results (the $\overline{IMC}(dB)$ improves almost 1 dB on the average).

The motion-compensated frames corresponding to the effect of additive noise on the synthetic sequence for SNR = 20 dB can be seen in Figure 2.

Table 2 shows the values for the MSE, bias, $\overline{IMC}(dB)$ and $\overline{DFD}^2$ for the estimated optical flow using the Wiener filter and the four programs mentioned previously with two noisy frames (SNR = 20 dB).

The results for both the noiseless and noisy cases present better values of $\overline{IMC}(dB)$ and $\overline{DFD}^2$ as well as MSE's and biases for all algorithms using the GCV. The best results in terms of metrics and visually speaking are obtained with the LSCRV2 algorithm ($\Lambda = \text{diag}\{\lambda_1, \lambda_2\}$ and multi-mask strategy). For the noisy case, it should be pointed out the considerable reduction of the interference of noise when it comes to the motion in the background and inside the object. For this algorithm, even the motion around the borders of the rectangle is clearer than when the LMMSE estimator is used.

**Experiment 2.** Figure 3 presents the values of the improvement in motion compensation for frames 31 to 40 of the MD sequence for the noiseless and noisy (SNR=20dB) cases, respectively, for all algorithms investigated. Here we concentrate our analysis on the performance of LSCRV2, which is the algorithm that gave us the best results. The LSCRV2 algorithm provides, on the average, 1.5 dB higher $\overline{IMC}(dB)$ than the LMMSE algorithm for the noiseless case. The $\overline{IMC}(dB)$ for the noisy case is not as high as in the previous situation. Their qualitative performance can be observed in Figure 4. By visual inspection, the noiseless case does present dramatic differences between both motion fields. For the noisy case, we were able of capturing the motion relative to the rotation of the mother's head, although incorrect displacement vectors were found in regions were there is no texture at all such as the background, for instance, but there is less noise than when we use the Wiener filter.

**Experiment 3.** Figures 5 and 6 demonstrate results obtained for frames 11-20 of the "Foreman" sequence. Some frames of this sequence show abrupt motion changes. One can see that all the algorithms based on

GCV outperform the LMMSE. This sequence shown very good values for the $\overline{IMC}(dB)$ for both the noiseless and the noisy cases. As one can see by looking at the plots for the errors in the motion compensated frames, the algorithm LSCRV2 performs better than the Wiener filter visually speaking.

## 8. Conclusions

First, it should be pointed out that our GCV model can handle the motion estimation/detection problem well and, as expected, $\Lambda$ gives better result than a scalar regularization parameter.

Second, the proposed GCV method provides an automatic, data-based selection of the regularization parameter by means of the minimization of (14).

However, the technique presents some drawbacks [8]. This technique works very well in most of the cases (approximately 95% of the time), but due to the volume of minimizations done, the GCV failed to produce good estimates at all points because of one of the following situations:

**a)** GVC($\Lambda$) has multiple minima.
**b)** There is no minimum such that all entries of $\Lambda$ are positive.
**c)** The minimum is hard to be found (no convergence).
**d)** The global minimum of the GCV results in a undersmoothed solution; a local minimum can be better.
**e)** We may have found a saddle point.

Our spatially adaptive scheme indeed improves the behavior of the routines based on GCV around motion borders due to the fact that it seeks the neighborhood which provides the best system of equations according to the smoothness constraint assumption.

An interesting problem we are currently investigating, is a more intelligent way of choosing a neighborhood upon which to build our system of equations. We are also looking at more complex regularization matrices.

Table 1. Comparison between GCV implementations and the Wiener filter. SNR = ∞.

|  | Wiener | LSCRV | LSCRVB | LSCRV1 | LSCRV2 |
|---|---|---|---|---|---|
| $MSE_x$ | 0.1548 | 0.1534 | 0.1511 | 0.1493 | 0.1440 |
| $MSE_y$ | 0.0740 | 0.0751 | 0.0753 | 0.0754 | 0.0754 |
| $bias_x$ | 0.0610 | 0.0619 | 0.0599 | 0.0581 | 0.0574 |
| $bias_y$ | -0.0294 | -0.0291 | -0.0294 | -0.0294 | -0.0293 |
| $\overline{IMC}(dB)$ | 19.46 | 19.62 | 19.74 | 19.89 | 20.38 |
| $\overline{DFD}^2$ | 4.16 | 4.05 | 3.921 | 3.76 | 3.35 |

Table 2. Comparison between GCV implementations and the Wiener filter. SNR = 20dB.

|  | Wiener | LSCRV | LSCRVB | LSCRV1 | LSCRV2 |
|---|---|---|---|---|---|
| $MSE_x$ | 0.2563 | 0.2544 | 0.2446 | 0.2437 | 0.2373 |
| $MSE_y$ | 0.1273 | 0.1270 | 0.1268 | 0.1257 | 0.1254 |
| $bias_x$ | 0.0908 | 0.0889 | 0.0883 | 0.0881 | 0.0852 |
| $bias_y$ | -0.0560 | -0.0565 | -0.0564 | -0.0561 | -0.0553 |
| $\overline{IMC}(dB)$ | 14.74 | 14.83 | 14.98 | 15.15 | 15.32 |
| $\overline{DFD}^2$ | 12.24 | 12.02 | 11.60 | 11.16 | 10.78 |

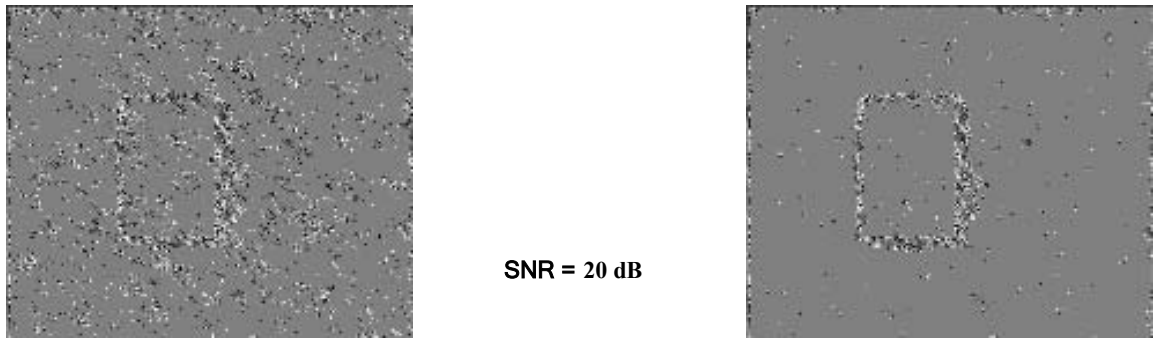

SNR = 20 dB

Figure 2. Motion-compensation errors for the LMMSE (left) and the LSCRV2 (right) algorithms.

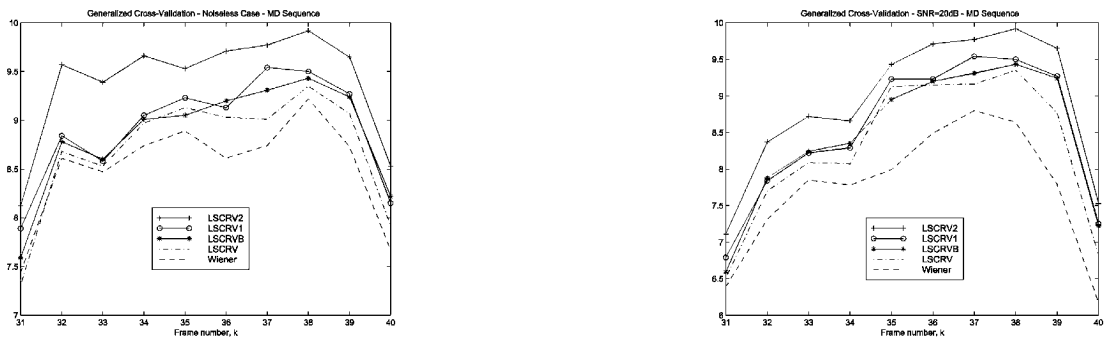

Figure 3. $\overline{IMC}(dB)$ for the noiseless (left) and noisy (right) cases for the MD sequence.

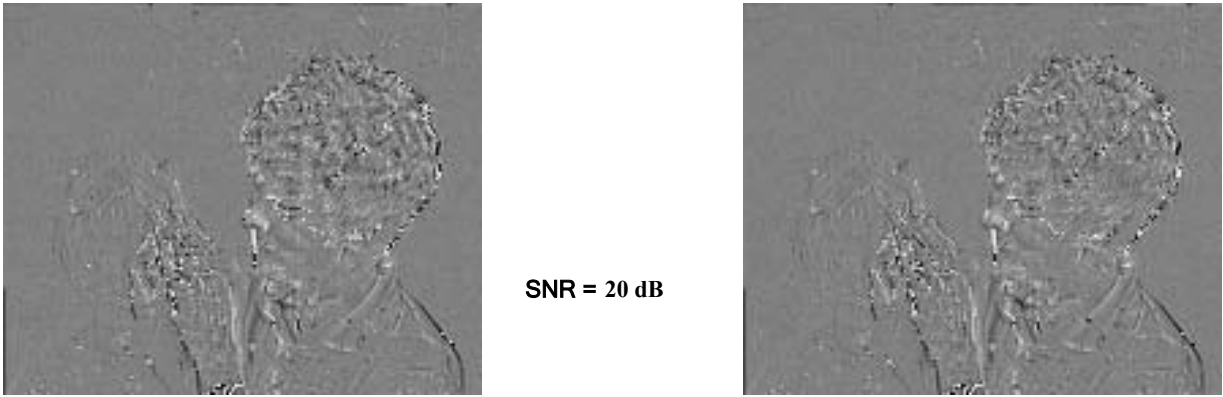

Figure 4. Motion-compensated errors for frame 32: the LMMSE (left) and the LSCRV2 (right) algorithms.

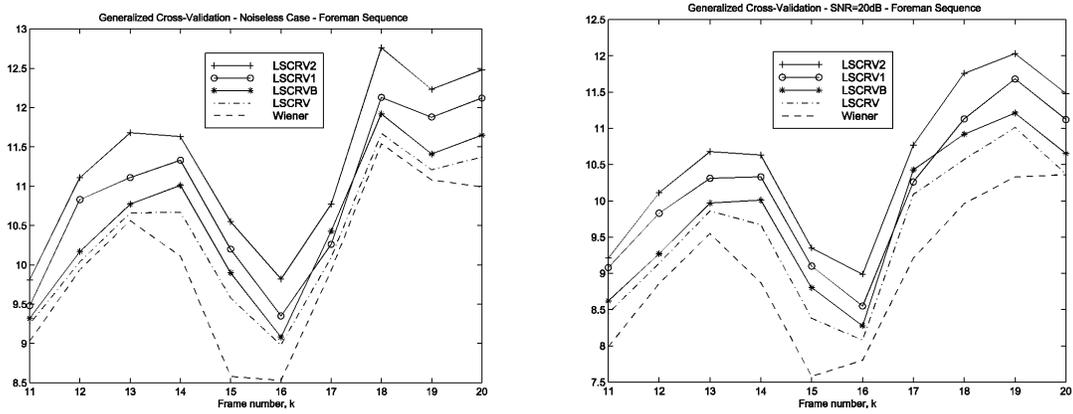

Figure 5. $\overline{IMC}(dB)$ for frames 11-20 of the noiseless (left) and noisy with SNR=20dB (right) for the Foreman sequence.

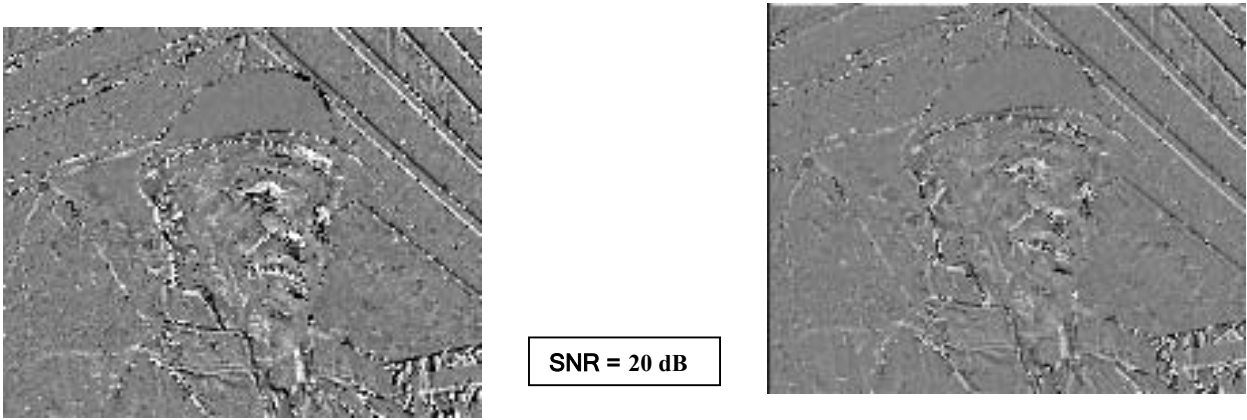

Figure 6. Motion-compensated errors for frame 16: the LMMSE (left) and the LSCRV2 (right) algorithms.